\documentclass{article}

\PassOptionsToPackage{numbers, compress}{natbib}

\usepackage[preprint]{neurips_2021}

\usepackage[utf8]{inputenc} 
\usepackage[T1]{fontenc}    
\usepackage{hyperref}       
\usepackage{url}            
\usepackage{booktabs}       
\usepackage{amsfonts}       
\usepackage{nicefrac}       
\usepackage{microtype}      
\usepackage{algorithm}
\usepackage{algorithmic}
\usepackage{wrapfig}
\usepackage{graphicx}
\usepackage{caption}
\usepackage{subcaption}
\usepackage{color}

\usepackage{latexsym}
\usepackage{amsmath}
\usepackage{amssymb}
\usepackage{amsthm}
\usepackage{url}
\usepackage{color}

\theoremstyle{definition}
\newtheorem{thm}{Theorem}

\newtheorem{prop}[thm]{Proposition}

\newcommand{\RR}{\mathbb{R}}

\newcommand{\mcG}{\mathcal{G}}

\newcommand{\mcS}{\mathcal{S}}

\newcommand{\mcZ}{\mathcal{Z}}

\newtheorem*{prop*}{Prposition}
\title{Contrastive Representation Learning with\\
Trainable Augmentation Channel}

\author{
  Masanori Koyama, Kentaro Minami, Takeru Miyato  \\
  Preferred Networks, Inc. \\
  Tokyo, Japan \\
  \texttt{\{masomatics, minami, miyato\}@preferred.jp} \\
  \And
  Yarin Gal \\ 
  University of Oxford \\
  Oxford, United Kingdom\\
  \texttt{yarin@cs.ox.ac.uk} 
}

\begin{document}

\maketitle

\begin{abstract}

In contrastive representation learning, data representation is trained so that it can classify the image instances even when the images are altered by augmentations.
However, depending on the datasets, some augmentations can damage the information of the images beyond recognition, and such augmentations can result in collapsed representations.
We present a partial solution to this problem by formalizing a stochastic encoding process in which there exist a tug-of-war between the data corruption introduced by the augmentations and the information preserved by the encoder.
We show that, with the infoMax objective based on this framework, we can learn a data-dependent distribution of augmentations to avoid the collapse of the representation.
\end{abstract}

\section{Introduction}\label{sec:introduction}



Contrastive representation learning (CRL)  is a family of methods that learns an encoding function $h$ so that, in the encoding space, any set of augmented images produced from a same image (positive samples) are made to attract with each other, while the augmented images of different origins(negative samples) are made to repel from each other \cite{hjelm2018learning, bachman2019amdim, henaff2020cpc, tian2019cmc, chen2020simclr}. 
Oftentimes, the augmentations used in CRL are chosen to be those that are believed to maintain the "content"\begin{footnote}{If $Y$ is a target signal, we may for  example assume $P(Y|T(X))=P(Y|X)$, as in \cite{Imsat}}\end{footnote} features of the inputs, while altering the "style" features to be possibly discarded in the encoding process \cite{content_isolate}. 
However, how can we be so sure that a heuristically chosen set of augmentations does not affect the features that are important in the downstream tasks?  
For example, consider applying a cropping augmentation $T$ to a dataset consisting of MNIST images located at random position in blank ambient space(Figure\ref{fig:regional}).
\begin{figure}[h!]
\begin{center}
\includegraphics[scale=0.39]{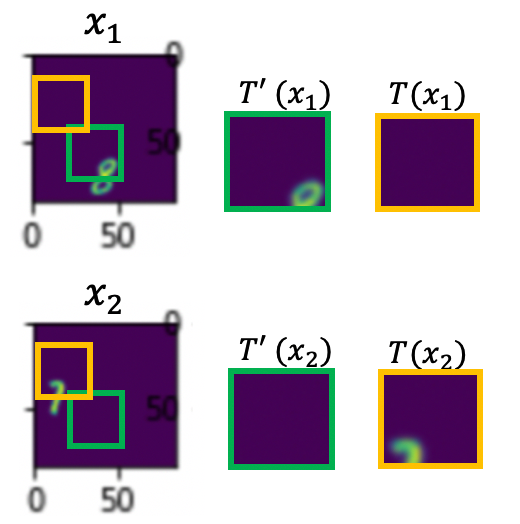}
\includegraphics[scale=0.43]{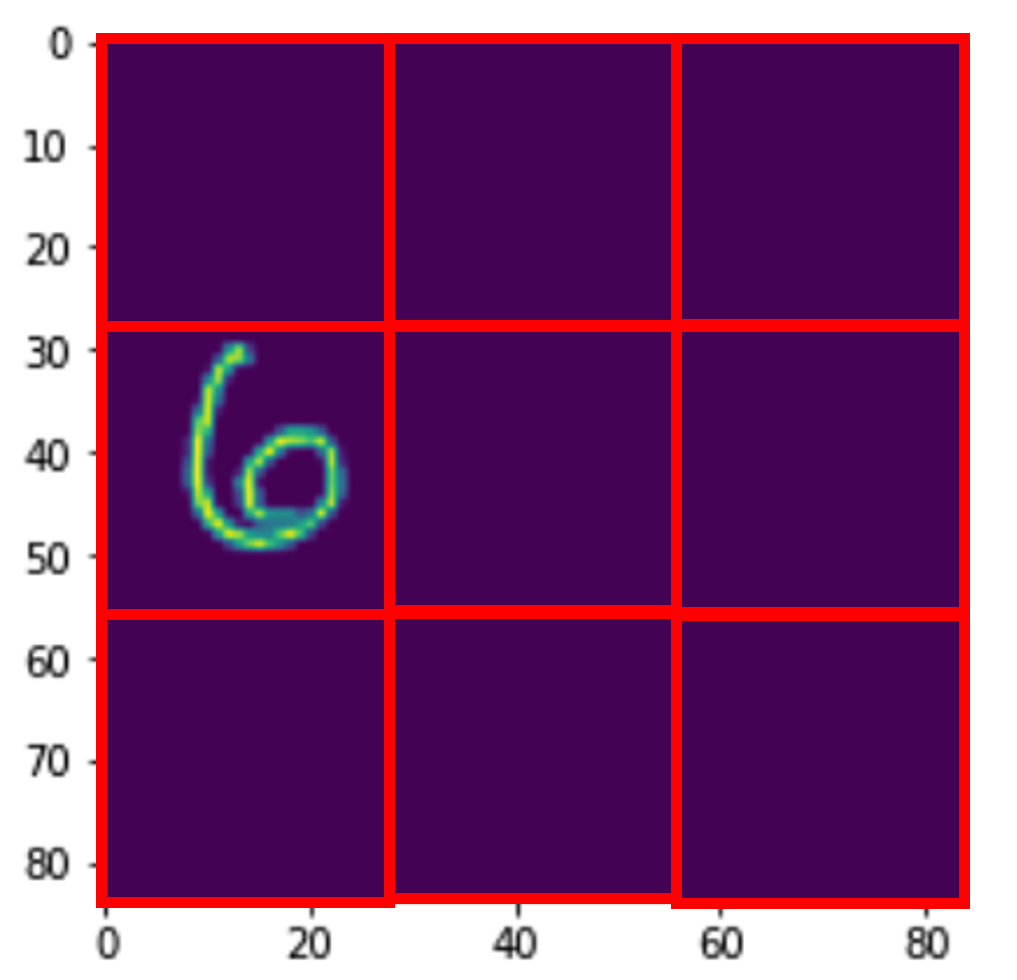}
\includegraphics[scale=0.43]{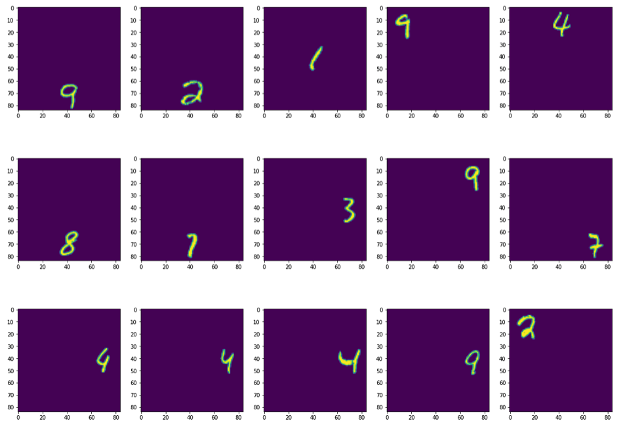}
\end{center}
\caption{
The leftmost Panel: If we enforce the equivalence relation $T(x_k) \sim T'(x_k)$, then we will also have $T(x_2) \sim T(x_1)$ by transitivity because $T'(x_2)=T(x_1)$.
Right two panels: Example images of the MNIST-derived dataset and 9 positions on which a MNIST digit was placed in each one of 
 $(28 * 3) \times (28 * 3)$ dimensional image.  } 
\end{figure} \label{fig:regional} 
In this case, since $T'(x_2)=T(x_1)$, training an encoder $h$ such that $h(T'(x_k)) \cong h(T(x_k))$ would also force $h(T(x_1)) \cong h(T(x_2))$ by the transitivity of "$\cong$".
In a semi-supervised setting, such a problem of \textit{wrong clustering} may be avoided by considering a stochastic $T$ with a distribution $P(T|X)$ satisfying $P(Y|T(X)) \cong P(Y|X)$, as in \cite{Imsat}. 

In our study, we provide a partial solution to this problem in a self-supervised setting.
In particular, we formalize the representation $Z$ as the output of a stochastic function parametrized by an encoder function $h$ and a stochastic augmentation $T$, and maximize $I(X;Z)$ in a tug-of-war between the data corruption introduced by $T$ and the information preserved by $h$. 
Although the infoMax in the context of $I(T(X), T'(X))$ has been discussed in previous literatures \citep{tian2019cmc,bachman2019amdim,tschannen2020mutual, MI_contra}, it has not been investigated thoroughly while giving a freedom to the distribution of $T$. 
We will empirically demonstrate that we can learn a competitive representation by training $P(T|X)$ together with $h$ in this framework.
Our formulation of $I(X;Z)$ also provides another way to interpret simCLR \cite{chen2020simclr} as a special case in which $P(T|X)$ is fixed to be the uniform distribution. 

\vspace{-0.5cm}

\section{InfoMax problem with Augmentataion Channel}\label{sec:methods}

\begin{figure}[ht!]
         \centering
         \includegraphics[scale=0.35]{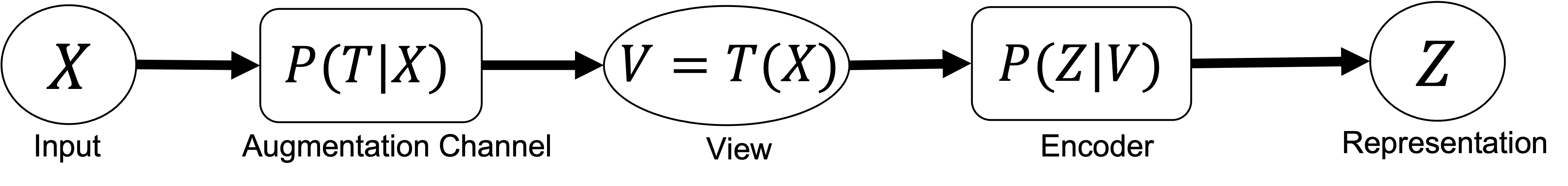}
         \caption{Generation Process of $Z$}
         \label{fig:non-branching}
\end{figure}

Existing perspectives of CLR are based on $I(T(X); T'(X))$ (discussed more in depth in related works, Section \ref{sec:related}).
In this work, we revisit the infoMax problem from a different perspective in a framework of self-supervised learning that explicitly separates the \textit{augmentation channel} in the encoding map $X \to Z$. 
Consider the generation process illustrated in the Figure  \ref{fig:non-branching}.
In this process, $V=T(X)$ is produced from $X$ by applying a random augmenation $T$ sampled from some distribution $p(T|X)$. 
$V$ is then encoded into $Z$ through the distribution $p(Z|V)$ parametrized by some encoder $h$.
Thus, the distribution of $Z$ can be written as
\begin{align}
    p(z| x) =  \int p\left(z | T(x)\right)p(T| X) dT    
\end{align}
Using $\mcG$ to denote the family of distributions that can be written in this form, we consider the InfoMax problem $\max_{p \in \mcG} I(X; Z)$.
In this definition of the map $X \to Z$, the support $\mathcal{T}$ of $p(T|X)$ determines the maximum amount of information that can be preserved.
For example, if all members of $\mathcal{T}$ strongly corrupts $X$, $I(X;Z)$ would be small for all choice of $P(T|X)$. 
Meanwhile, if the identity transformation is included in $\mathcal{T}$, then $V=X$ can be achieved by setting $P(T|X)=\delta_{id}(T)$.
However, as in training methods based on noise regularization \cite{VAT, Noise_reg,Imsat}, the identity mapping is often not included in the augmentation set because it does not help regularize the model. 

The infoMax problem in our framework has a deep connection with modern self supervised learning, as it can provide another derivation of simCLR that does not use a variational approximation.
\begin{prop}

Suppose that $p(Z \mid T(X))= C_\beta \exp(\beta \mcS(Z, h(T(X))))$      
where $\mcS: \mcZ \times \mcZ \to \RR$ is a similarity function on the range of $Z$ and  $C_\beta$ is a constant dependent only on $\beta$. Then
\begin{align}
 I(X ; Z) &=  E_{X, Z} \left[ \log  E_{T'|X}\left[ \frac{  \exp(\beta \mcS(Z , h(T'(X))) }{E_{T'', \tilde X}[\exp(\beta \mcS(Z, h(T'' (\tilde X)))]}\right]  \right] \label{eq:TBCRL} 
\end{align} 
Also, when $P(T|X)$ is uniformly distributed on a compact set of view-transformations,  the mean approximation of $Z$ and Jensen's inequality on the $E_{T'|X}$  part of \eqref{eq:TBCRL} recovers the simCLR loss.
\label{thm:TBC_prop}
\end{prop} 
For the proof of Prop \ref{thm:TBC_prop}, please see Appendix \ref{appsec:proof}. We shall note that the condition of this statement is fulfilled in natural cases, such as when $P(Z|T(X))$ is Gaussian or Gaussian on the sphere. 
In the proof of Prop \ref{thm:TBC_prop}, the numerator and the denominator correspond directly to $- H(Z | X)$ and $H(Z)$.
If $Z$ takes its value on the sphere $\mcS^{d}$, enlarging $H(Z)$ would encourage $Z$ to be uniformly distributed over the sphere. These observations support the theory proposed in  \cite{wang2020uniform}. 
The table in Appendix \ref{appsec:algorithm} summarizes our algorithm for optimizing the objective \eqref{eq:TBCRL} with respect to both $P(T|X)$ and $h$.

\vspace{-0.15cm}

\section{Experiments}\label{sec:experiments}
\vspace{-0.15cm}

We show that, by training $P(T|X)$ together with the encoder $h$ based on the objective \eqref{eq:TBCRL}, we can learn a better representation than the original simCLR.
We conducted an experiment on a dataset derived from MNIST mentioned at the introduction (Figure\ref{fig:regional}).
To construct this dataset, we first prepared a blank image of size $(28 * 3) \times (28 * 3)$, which is $3$ times greater in both dimensions than the original MNIST images ($28 \times 28$).  
We then created our dataset by placing each MNIST image randomly at one of $3 \times 3 = 9$ grid locations in the aforementioned blank image.
We set $T$ to be a random augmentation that crops a $20 \times 20$ image at one of $17 \times 17=289$ locations ranging over the $(28 * 3) \times (28 * 3)$ dimensional image with stride size $4$. 
On this dataset, any crop that does not intersect with the digit produces the same \emph{empty} image, which is useless in discriminating the image instances.
For computational ease, we trained our encoder $h$ based on the Jensen-lower bound of \eqref{eq:TBCRL}.
We shall also note that, in our setup, our $h$ corresponds to the composition of the projection head $g$ and the encoder $f$ in the context of the recent works of contrastive learning.
We evaluated the representation of both $h = g \circ f$ and $f$.  
Also, without any additional constraint, $P(T|X)$ sometimes collapsed to the "the most discriminating" crop on the training set, resulting in a representation that does not generalize on the downstream classification task. 
To resolve this problem, we adopted the maximum entropy principle \cite{softQ_Haarnoja} and 
optimized our objective \eqref{eq:TBCRL} together with small entropy regularization $H(T|X)$, seeking the highest entropy $T$ that maximises the objective \eqref{eq:TBCRL}.


\vspace{-0.15cm}

\subsection{Performance of the trained representations in Linear Evaluation Protocol} \label{sec:exp_table1}
To evaluate the learned representation, we followed the linear evaluation protocol as in \cite{chen2020simclr} and trained a multinomial logistic regression classifier  on the features extracted from the frozen pretrained network. We used Sklearn library \cite{pedregosa2011scikit} to train the classifier. 
For SimCLR, it is often customary to use the "center crop" augmentation $T_{center}$ and report $h(T_{center}(X))$ as the representation for $X$. However, in this example, "center crop" would extract an empty image with high probability. 
Thus, we computed the representation of each $X$ by integrating the encoded variable with respect to $P(T|X)$, that is, $\hat{Z} = E_{T\sim P(T|X)}[h(T(X))]$ ($P(T|X)$ for simCLR is uniform). 
For the models with non-uniform $P(T|X)$ we also evaluated $Z_{topn}$, the representation obtained by averaging $h(T(X))$ over the set of $T$s having the top eight $P(T|X)$ density. 
As an ablation, we also evaluated the SimCLR-trained encoder by integrating its output with respect to the oracle $P(T|X)$ concentrated uniformly on the $9$ crop positions with maximal intersection with the embedded MNIST image. We conducted each experiment with $4$ seeds. 
The table \ref{tab:lcp} summarizes the result. 
\begin{table}[t]
\small
  \caption{Linear evaluation accuracy scores. Raw Representation achieves $0.8992 \pm 0.0012$. For the description of \textit{oracle} and \textit{topn}, please see the main script (Section \ref{sec:exp_table1}).}
  \label{tab:lcp}
  \centering
  \scalebox{0.9}{
  \begin{tabular}{lccccc}
    \toprule
    \toprule
    Method  &  Ours  &  Ours(topn)  &  SimCLR  & simCLR(oracle)  \\
    \midrule      
    Projection Head  & $0.95505 \pm 0.0023$ & $0.9552 \pm 0.0037$ & $0.3156 \pm 0.0044$ & $0.5144 \pm 0.011$ \\ 
    $f$ output & $0.9729\pm 0.0014$ & $0.9748 \pm 0.0012$ & $0.4598\pm 0.0056$ &  $0.9354 \pm 0.0029$
\\
    \bottomrule
 \end{tabular}}
\end{table}

We can see that, with our trained $P(T|X)$ and $h$, we can achieve a very high linear evaluation score, even better than the raw representation result on the ordinary MNIST dataset ($0.9256$). 
Interestingly, with our $P(T|X)$, the representation is competitive even at the projection head, and its performance even exceeds the representation of simCLR obtained by averaging $f(T(X))$ over the oracle $P(T|X)$.  
This trend was also observed in the experiment on the original MNIST(see Appendix \ref{appsec:mnist}).
This result may suggest that the poor quality of simCLR representation at the level of the projection head is partially due to the fact that proper $P(T|X)$ is not used in training the model. 
Also, in confirmation of our problem statement in the section \ref{sec:introduction}, the representation learned without the trainable $P(T|X)$ collapses around that of the empty image (see Appendix \ref{appsec:uniformity}). In terms of the average pairwise Gaussian potential used in \cite{wang2020uniform} that measures the uniformity of the representations on the sphere(lower the better), our representation achieves $0.0845$ as opposed to $0.9757$ of the baseline simCLR.  

\vspace{-0.15cm}

\subsection{The trained $P(T|X)$ agrees with our intuition} 
\begin{figure}[h!]
\begin{center}
\includegraphics[scale=0.3]{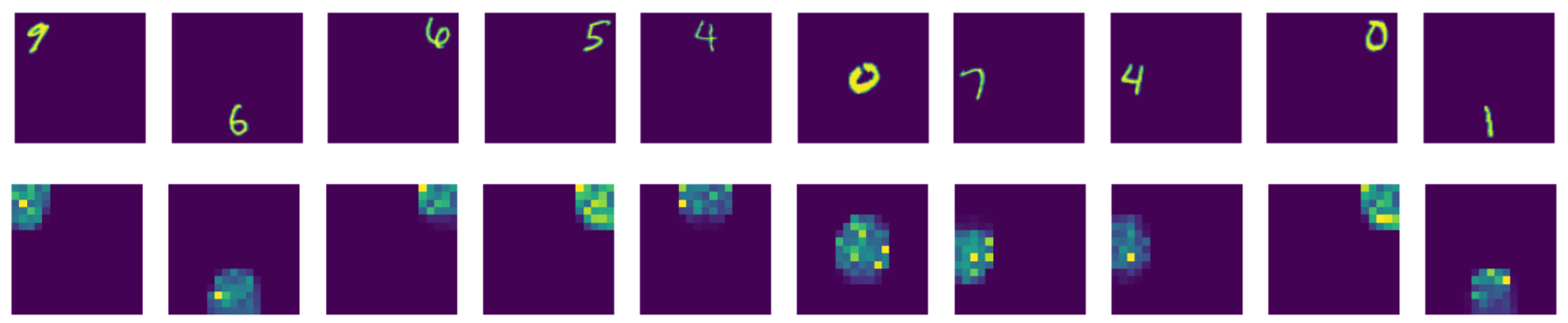}
\caption{Visualization of the trained $P(T|x)$ (bottom row) for various choice of $x$(top row). Brighter color represents higher intensity. } 
\label{fig:ptgivenx} 
\end{center}  
\end{figure}
Figure 3 visualizes the density of $P(T|x)$(second row) for various input image $x$(second row). In each image of the second row, the intensity at $(i,j)$th pixel is $P(T_{ij}|X)$, where $T_{ij}$ is the augmentation that crops the sub-image of size $20 \times 20$ with the top left corner located at $(i,j)$.
As we can see in the figure, the learned $P(T|X)$ is concentrated on the place of digit, ignoring the crop locations that would return the empty image. 
Our learned $P(T|X)$ in fact captures the non-trivial crop with probability $0.998 \pm 0.003$ on 10,000 test images. 

\vspace{-0.15cm}

\section{Related Works and conclusion}\label{sec:related}
\vspace{-0.15cm}

In a way, $P(T|X)$ can be considered an augmentation \emph{policy}.
\citep{AutoAug, FAA} also learns $P(T|X)$ with supervision signals. 
\citep{SelfAutoAug} extends these works to self-supervised setting
by applying a modified \citep{FAA} to a set of self-supervised tasks that are empirically 
 correlated to the target downstream tasks. 

There also are several works that investigate the importance of non-uniform sampling in the constrative learning. 
For example,  \cite{Good_view} proposes the infoMin principle, which claims that one shall engineer the distribution of $T(X)$ in such a way that it (1) shares as much information as possible with the target variable $Y$ while (2) ensuring that, for any two realization $t_1\neq t_2$ of $T$, $t_1(X)$ and $t_2(X)$ should have as little information in common. 
In their work, however, they do not provide an algorithm to optimize the distribution of $T$.
In a way, the requirements (1) and (2) seem to be respectively related to $H(X|Z)$ and $H(Z)$  in the numerator-denominator decomposition of \eqref{eq:TBCRL}.  
Also, because they are practically conducting an empirical study on the joint distribution $P(T_1, T_2)$,
their work might be also related to the optimization of $P(Z|T(X))$ in our context. 
Also, \cite{viewmaker} trains $T$ adversarially with respect to the loss. However, in the setting we discuss in this paper, this strategy would encourage $T$ to crop only the empty image and collapse the representations.



Previously, the connection between CLR and Mutual information has also been described based on the perspective that interprets CLR as a variational approximation of the mutual information between two views $I(V_1; V_2)$, where each $V_k = T_k(X)$ is a "view" of $X$ produced by some augmentation function $T_k$ \citep{tian2019cmc,bachman2019amdim,tschannen2020mutual, MI_contra}. 
This variational approximation is based on the inequality
\begin{align}
\begin{split}
    I(V_1; V_2) \geq E_{V_1, V_2} \left[ \frac{\exp(f(V_1,V_2))}{E_{ V'_1}[\exp(f(V'_1,V_2)]} \right]
\end{split}\label{eq:InfoNCE}
\end{align}  
that holds for \textit{any} measurable $f$. 
Based on this infoNCE perspective, \cite{tschannen2020mutual} considers a case in which $Z$ is trained as $Z = g(V)$ with invertible $g$,  
and presents an empirical study suggesting that simCLR can improve the representation even in this setting.  Based on this argument, \cite{tschannen2020mutual} suggests that $I(Z_1, Z_2)$ cannot be used to explain the success of simCLR.
However, as we point in our study, the transformation $X \to V$ usually involves information \emph{loss} via augmentations like \textit{cropping}, and CLR is often evaluated based on $Z$ sampled from $P(Z|V)$. 
In this study, we formalize the augmentation channel $X \to V$ as a part of $X \to Z$,  and present a result suggesting that, at least for the learning of $P(T|X)$, the Mutual information (MI) with $H(T|X)$ regularization might be an empirically useful measure for learning a good representation, in particular at the level of final ouput(projection head).  
Our result may suggest that it might be still early to throw away the idea of MI in all aspects of the CLR because \cite{tschannen2020mutual} studies a case in which only the $V\to Z$ part of $X \to Z$ is made invertible. 

It might also be worthwhile to mention some theoretical advantages of our formulation. 
Because \eqref{eq:InfoNCE} is a variational bound that holds for any choice of $f$,  this inequality does not help in estimating how much the RHS derived from a \emph{specific} choice of $f$ (i.e. RHS(f)) differs from $I(V_1; V_2)$.
Also, when we optimize RHS($f$) using a popular family of $f$ defined as $f(V_1, V_2) :=  \psi(h(V_1))^T \psi(h(V_2))$ \cite{MI_contra}, there is no way to know "in what proportion a given update of $f$ would affect $I(h(V_1); h(V_2))$ and $I(V_1; V_2) - RHS(f)$.   
Meanwhile, in our formulation, the difference between simCLR and MI is described directly with Jensen and mean approximation, for which there are known mathemtical tools like \cite{JensenGap}.
It might be interesting to further investigate the claims made by \cite{tschannen2020mutual} in this direction as well.
We believe that our approach provides a new perspective to the study of contrastive learning as well as insights to the choice of augmentations.



\vspace{-0.15cm}

\bibliographystyle{plainnat}
\bibliography{references}
\vspace{15cm}

\newpage
\section{Appendix} 
\subsection{Formal statement and the proof of Proposition \ref{thm:TBC_prop}} \label{appsec:proof}

\begin{prop*}

Suppose that $p(Z \mid T(X))= C_\beta \exp(\beta \mcS(Z, h(T(X))))$      
where $\mcS: \mcZ \times \mcZ \to \RR$ is a similarity function and  $C_\beta$ is a normalization constant dependent only on $\beta$,
Then 
\begin{align}
 I(X ; Z) &=  E_{X, Z} \left[ \log  E_{T'|X}\left[ \frac{ \exp(\beta \mcS(Z , h(T'(X)))}{E_{T'', X'}[\exp(\beta \mcS(Z, h(T'' (X')))]}\right] \right].
\end{align} 
Also, when $P(T|X)$ is uniformly distributed over a compact set of view-transformations, we recover the loss of SimCLR by 
(1) applying Jensen's inequality on $E_{T'|X}$ and 
(2) approximating $Z$ with $h(T(X))$, the mean of $p(Z|T(X))$. 
\end{prop*}

\begin{proof}
We use upper case letter to denote the random variable and lower case letter to denote its corresponding realization ($x$ is a realization of $X$).
We also use the standard notation in the measure theoretic probability that treat expressions like $P(A| B)$ and $E[A| B]:=E_{A|B}[A]$ as a random variable that is measurable with respect to $B$.
Thus, in the equality $E[A] = E[E[A | B]]$, the integral $E[A | B]$ inside the RHS is a random variable with respect to $B$.  
To clarify, we sometimes use the subscript to represent the variable with respect to which the integral is taken. 
For more details about this algebra, see \cite{Durrett} for example. 
Here, we show the proof of the version of the statement with the application of Jensen's inequality. 
The proof without Jensen's inequality can be derived easily from the intermediate results of this proof.


\paragraph{On $-H(Z \mid X)$}
\begin{align}
    E_{X,Z}[\log P(Z|X)] &=  E_{X,Z}[\log ( E_{T'}[P(Z|X, T') | X] )] \\
    &:= E_{X,Z}[ \log E_{T'|X}[ (C_\beta \exp(\beta S(Z, h( T'(X))))) ] \\
    &=E_{X,Z}[ \log E_{T'|X}[ (\exp(\beta S(Z, h(T'(X))))) ] + C_\beta  \\
    &\geq E_{X,Z}[ E_{T'|X}[\log (\exp(\beta S(Z, h(T'(X)))))] + C_\beta \\
    &= E_{X,Z}[ E_{T'|X}[\beta S(Z, h(T'(X)))]  + C_\beta  \\
    &:= E_{X,Z} [E_{T'|X}[\beta S(Z, h(T'(X)))] + C_\beta
\end{align}

\paragraph{On $H(Z)$}
\begin{align}
    - E[\log P(Z)] &= -E_Z[\log (E_{X', T''}[P(Z|X', T'')])] \\
    &= -E_Z[\log (E_{X', T''} [C_\beta \exp(\beta S(Z, h(T''( X')))) ])] \\
    &= - E_Z[\log (E_{X', T''} [\exp(\beta S(Z, h(T''( X')))) ])] - C_\beta
\end{align}

Altogether, we see that $C_\beta$ cancels out and 
\begin{align}
    H(Z) - H(Z \mid X) &\geq E_{X,Z} [E_{T'|X}[\beta S(Z, h(T'(X)))] + C_\beta \\
    & ~~~~ - \log (E_{X', T''} [exp(\beta S(Z, h(T''( X')))) ])] - C_\beta \\
    &= E_{X,Z}\left[ E_{T'|X}\left[ \log \frac{\exp(\beta S(Z, h(T'(X))))}{E_{X', T''} [\exp(\beta S(Z, h(T'( X')))) ]} \right] \right] \label{eq:inequality}  
\end{align}
The equality emerges if we do not apply Jensen's inequality on $-H(Z|X)$. 

To show the connection of this result with simCLR, we approximate $Z|X$ as $h(T(X))$, the mean of $P(Z|T(X))$.
With this approximation, the outermost integration with respect to $(X,Z)$ will be replaced by the integration with respect to $(X,T)$. 
Also, because $T''$ is integrated away in the denominator of \eqref{eq:inequality}, the \textit{double prime} superscript of the $T''$ is superficial. 
Thus, we obtain 
\begin{align}
    &E_{X,T}\left[ E_{T'|X}\left[ \log \frac{\exp(\beta S(h(T(X)), h(T'(X))))}{E_{X', T'} [\exp(\beta S(h(T(X)), h(T'( X')))) ]} \right] \right] \\
    &\cong \frac{1}{N}  \sum_{x_i \sim X, T_i \sim (T|x_i)} \Bigg( \frac{1}{\tilde N}  \sum_{T'_k \sim (T|x_i)}\beta S(h(T_i(x_i)), h(T'_k(x_i)))  \\
    - & \frac{1}{M} \log \left( \sum  \sum_{x_j \sim X, T'_j \sim (T|x_j)} \exp\left( \beta S(h(T_i(x_i)), h(T'_j(x_j))) \right) \right)    \Bigg)
\end{align}
With $i \in {1:N},  k \in {1:\tilde N}, j \in {1 : M}$. 

Choosing $\tilde N =1$ and $M=N$, we get 
\begin{align}
    \frac{1}{N}  \sum_{x_i \sim X, T \sim (T|x_i), T'_i \sim (T|x_i)} \log \Bigg( \frac{\exp( \beta S(h(T_i(x_i)), h(T'_i(X_i)))}{\frac{1}{N} \log \left( \sum_{x_j \sim X, T'_j \sim (T|x_j)} \exp\left( \beta S(h(T_i(x_i)), h(T'_j(x_j))) \right) \right)} 
    \Bigg)
\end{align}
which agrees with the simCLR loss when $T|X$ is set to be uniform.


\end{proof}

\subsection{Algorithm} \label{appsec:algorithm}
The table shown below is the description of the algorithm based on Proposition \ref{thm:TBC_prop} that trains $h$ and $P(T|X)$ together. 
In this algorithm we assume that the support of $P(T|X)$ is discrete.
Instead of training $h$ and $P(T|X)$ simultaneously, we train $h$ and $P(T|X)$ in turn because this strategy was able to produce more stable results.
With this algorithm's notation, the very classic SimCLR would emerge if we set $m$(the number of $T$ samples) to be $2$ and set $P(T|X)$ to be uniform.   
In our experiments we set $m$ to be $8$, as it performed better than anything less for both fixed $P(T|X)$(SimCLR) and trainable $P(T|X)$.  
\begin{algorithm}[ht!]
  \begin{algorithmic}[1]
    \renewcommand{\algorithmicensure}{\textbf{Input:}}  
    \REQUIRE A batch of samples $\{x_k\}$, an encoder model $h_\theta : x \to z$,
    the number of transformation samples $m$, a model for conditional random augmentation distribution $x \to P(T | x, \eta)$
     \FOR {each iteration $i$}
        \STATE \textbf{Update phase for $h$}
        \STATE Sample $T_{jk} \sim P(T | x_k, \eta)$, $j=1,...,m$ 
        \STATE Apply $\{ T_{jk} ; j = 1,...,m\}$ to each $x_k$, producing a total of
        $m \times k$ samples of $T_{jk}(x_k)$. 
        \STATE Empirically compute the objective  \eqref{eq:TBCRL} or its lower bound, and update $\theta$ 
        \STATE \textbf{Update phase for $P(T|X)$}
        \STATE Sample $T_{jk} \sim Uniform$
        \STATE Evaluate \eqref{eq:TBCRL} with $P(t_j | x_k, \eta)$ weights, and update $\eta$ 
     \ENDFOR
  \end{algorithmic} 
  \caption{Contrastive Representation learning with trainable augmentation Channel(CRL-TAC) }  \label{alg:algorithm}
\end{algorithm}  

\subsection{Model Architecture and entropy regularization }
\begin{figure}[h!]
    \centering
    \includegraphics[scale=0.4]{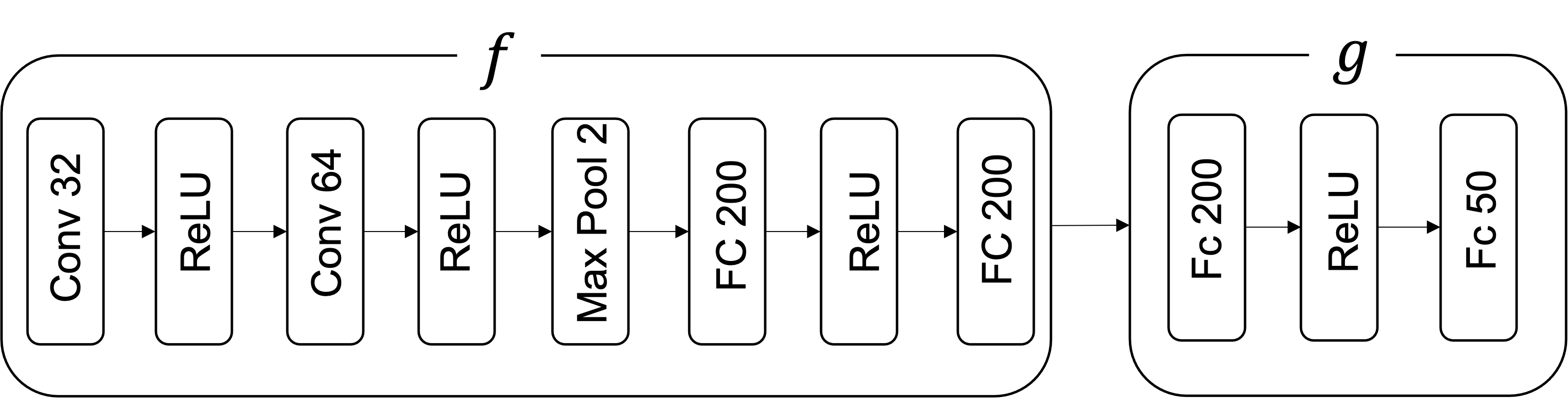}
    \caption{Encoder architecture}
    \label{fig:encoder_model}
\end{figure}

\begin{figure}[h!]
    \centering
    \includegraphics[scale=0.4]{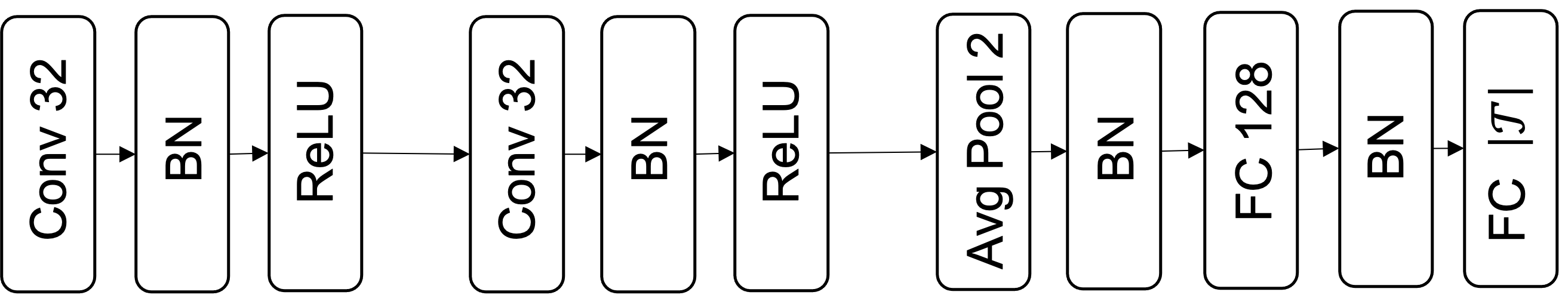}
    \caption{$P(T|X)$ architecture}
    \label{fig:ptgivenx_model}
\end{figure}

In our experiment, we used a three layer CNN with $200$ dimensional output for the intermediate encoder $f$ and a two layer MLP with $50$ dimensional output for the projection head $g$(Figure \ref{fig:encoder_model}).
We chose this architecture because this choice performed stably for SimCLR on standard MNIST dataset (See Section \ref{appsec:mnist}). 
We trained $P(T|X)$ with three layer CNN(Figure \ref{fig:ptgivenx_model}).

As in \cite{wang2020uniform}, we normalized the final output of the encoder $h = f \circ g$ so that the final output is distributed on the sphere.
As such, we used $S(a,b) = a^T b$, and set $\beta=0.5$ since this choice yielded stable results for the learning of $P(T|X)$.
At the inference time, we normalized $E_{P(T|X)} [h((T(X))]$.
To discourage $P(T|X)$ from collapsing prematurely, we imposed a regularization of $H(T|X)$ with coefficient $\lambda$. 
We used coefficient $\lambda =0.0025$, as it achieved the lowest contrastive loss on the training set in the range $[0.001, 0.005, 0.0025]$. 

This choice of $\lambda$ also produced the best linear evaluation score on the training dataset.
Setting $\lambda < 0.0001$ seemed to collapse $P(T|X)$ in many cases. 

\subsection{Results on the original MNIST dataset} \label{appsec:mnist}

Table \ref{tab:lcp_mnist} shows the results on the original MNIST dataset. 
We used the same setting as for the main experiment in Section \ref{sec:experiments}, except that we set $\beta = 1.0$.
On this dataset, raw representation achieves $0.9255$. 
When trained with uniform $P(T|X)$, the projection head representation is not much better than the raw representation. 
However, when trained together with $P(T|X)$, the projection head representation is comparable to the $f$ output. 
This result also suggest that, by training $P(T|X)$ together with $h=g \circ f$, we can improve the utility of the representation at the level on which the objective function function is trained, instead of the heuristically chosen intermediate representation $f$. 
This result also suggests that there is much room left for the study of the stochastic augmentation and intermediate representation.
\begin{table}[th!]
\small
  \caption{Linear evaluation accuracy Scores on the original MNIST dataset. Raw Representation achieves $0.9255 \pm 0.0001$ on the original MNIST dataset.}
  \label{tab:lcp_mnist}
  \centering
  \scalebox{0.9}{
  \begin{tabular}{lccccc}
    \toprule
    \toprule
    Method  &  ours  &  ours(topn)  &  SimCLR   \\
    \midrule      
    Projection Head  & $0.9642 \pm 0.0025$ & $0.9674 \pm 0.0015$ & $0.9273 \pm 0.0044$  \\ 
    $f$ output &  $0.9805 \pm 0.0006$ & $0.9859 \pm 0.0004$ & $0.9806 \pm 0.0056$
\\
    \bottomrule
 \end{tabular}}

\end{table}

\subsection{Uniformity of the learned representation} \label{appsec:uniformity}

\begin{figure}[h!]
\begin{center}
\includegraphics[scale=0.50]{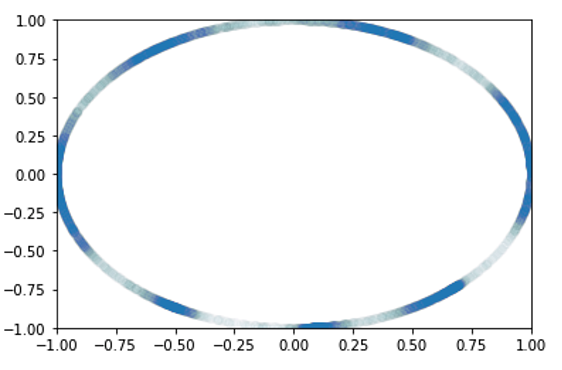}
\includegraphics[scale=0.5]{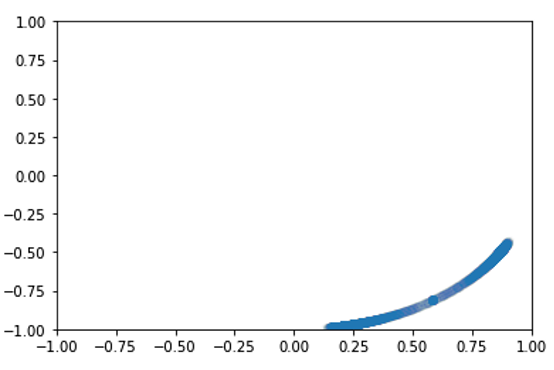}
\end{center} 
\caption{Left: The scatter plot of 2 dimensional representations trained together with $P(T|X)$. Right: The scatter plot of 2 dimensional representations trained with uniform $P(T|X)$.} 
\label{fig:2dim}
\end{figure}

\begin{figure}[h!]
\begin{center}
\includegraphics[scale=0.385]{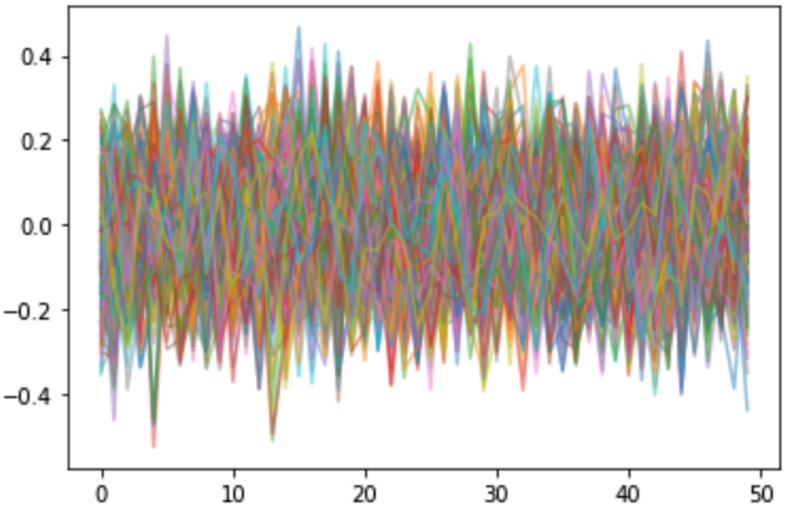}
\includegraphics[scale=0.4]{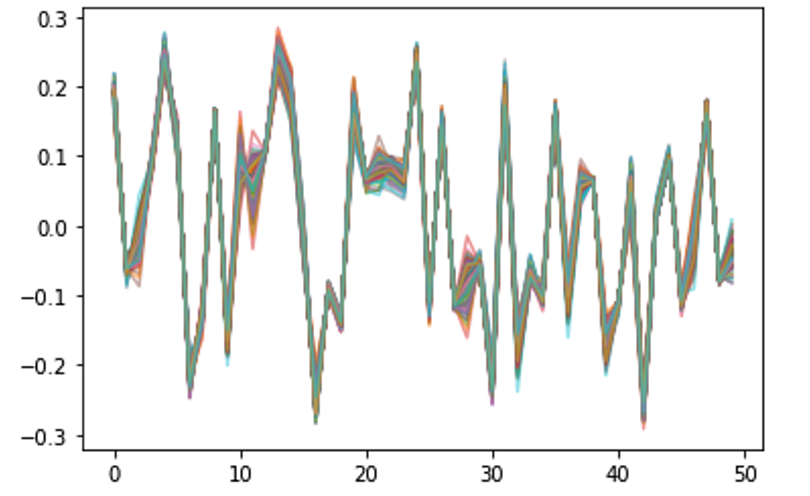}
\caption{Left: The superimposed plot of randomly sampled 200 instances of 50 dimensional representations trained together with $P(T|X)$. 
The horizontal axis represents the indices of the vectors, and each curve with a different color represents one instance of the vector $h(x) \in \mathcal{R}^{50}$. Right: The superimposed plot of 50 dimensional representations trained with uniform $P(T|X)$. We see that all instances of $h(x)$ look very similar.} 
\label{fig:50dim}
\end{center}
\end{figure}

\begin{figure}[h!]
\begin{center}
\includegraphics[scale=0.5]{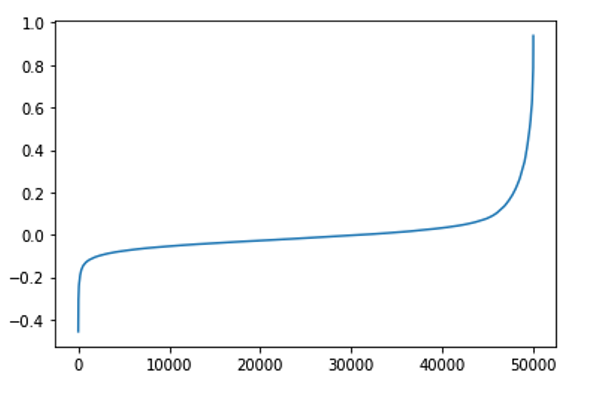}
\includegraphics[scale=0.5]{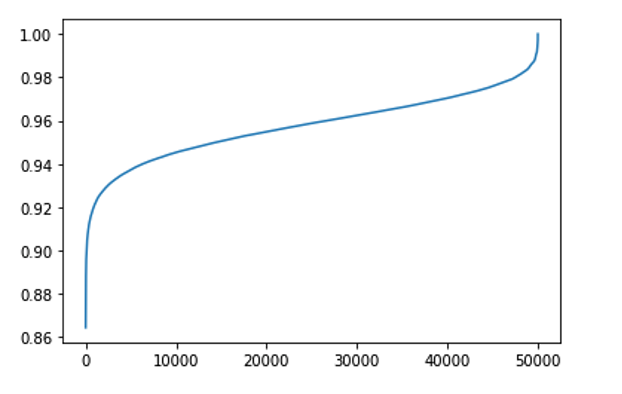}
\caption{Left : The plot of the sorted values of $|\langle h(x), h(x) \rangle|$ for a randomly sampled sets of $(x,x')$ pairs, when each $h(x)$ is a 50 dimensional representation trained together with $P(T|X)$.
Right: The same figure with $h$ trained with uniform $P(T|X)$.} 
\label{fig:prod}
\end{center}
\end{figure}

\cite{wang2020uniform} reports that, for a good representation, the representation tends to be more uniformly distributed on the sphere.  The graphs in Figure \ref{fig:2dim} are scatter plots of 2-dimensional representations trained with and without the trainable $P(T|X)$.
The graphs in Figure \ref{fig:50dim} are superimposed plots of 50 dimensional representations with and without the trainable $P(T|X)$.
On these graphs, we can visually see that what we feared in Section \ref{sec:introduction} and Figure \ref{fig:regional} happens when we fix $P(T|X)$; the majority of the representations becomes strongly concentrated around that of the empty image. 
This problem is successfully avoided with the trainable $P(T|X)$. 
In terms of the average pairwise Gaussian potential used in \cite{wang2020uniform} that measures the uniformity of the representations on the sphere(lower the better), our $50$ dimensional representation achieves $0.0845$ as opposed to $0.9757$ of the baseline SimCLR with fixed $P(T|X)$.
The graphs in Figure \ref{fig:prod} are the sorted values of $|\langle h(x), h(x') \rangle|$ for a randomly sampled set of $(x, x')$ pairs. 
We see in these graphs that the representations with the trainable $P(T|X)$ are trained to be as orthogonal to each other as possible($|\langle h(x), h(x') \rangle|$ is concentrated around $0$) , while the representations trained with the fixed $P(T|X)$ are collapsing into one direction ($|\langle h(x), h(x') \rangle|$ is concentrated around $1$).

\end{document}